\documentclass[10pt,twocolumn,letterpaper]{article}

\usepackage{cvpr}
\usepackage{times}
\usepackage{epsfig}
\usepackage{graphicx}
\usepackage{amsmath}
\usepackage{amssymb}

\usepackage[pagebackref=true,breaklinks=true,letterpaper=true,colorlinks,bookmarks=false]{hyperref}

\cvprfinalcopy 

\def\cvprPaperID{11} 

\ifcvprfinal\fi
\begin{document}

\title{Investigation on Combining 3D Convolution of Image Data and \\Optical Flow to Generate Temporal Action Proposals}

\author{Patrick Schlosser \\
Fraunhofer IOSB\\
\and
David M\"unch\\
Fraunhofer IOSB\\
{\tt\small david.muench@iosb.fraunhofer.de}
\and Michael Arens\\
Fraunhofer IOSB\\
}

\maketitle

\begin{abstract}
In this paper, several variants of two-stream architectures for temporal action proposal generation in long, untrimmed videos are presented. Inspired by the recent advances in the field of human action recognition utilizing 3D convolutions in combination with two-stream networks and based on the Single-Stream Temporal Action Proposals (SST) architecture \cite{sst}, four different two-stream architectures utilizing sequences of images on one stream and sequences of images of optical flow on the other stream are subsequently investigated. The four architectures fuse the two separate streams at different depths in the model; for each of them, a broad range of parameters is investigated systematically as well as an optimal parametrization is empirically determined. The experiments on the THUMOS'14 \cite{THUMOS14} dataset show that all four two-stream architectures are able to outperform the original single-stream SST and achieve state of the art results. Additional experiments revealed that the improvements are not restricted to a single method of calculating optical flow by exchanging the formerly used method of Brox \cite{brox2004high} with FlowNet2 \cite{ilg2017flownet} and still achieving improvements.
\end{abstract}


\section{Introduction}

Computer vision plays a major role in sports, beginning with an automatic semantic annotation of the observed scene to enhanced viewing experience. 

One major research field in computer vision is the recognition of actions and activities in videos, with a special interest in actions and activities performed by humans. The recognition of specific actions usually takes place in videos of limited length, called trimmed videos, by assigning a single action class to each video.  

\begin{figure}[t]
\begin{center}
\fbox{\includegraphics[width=.98\linewidth]{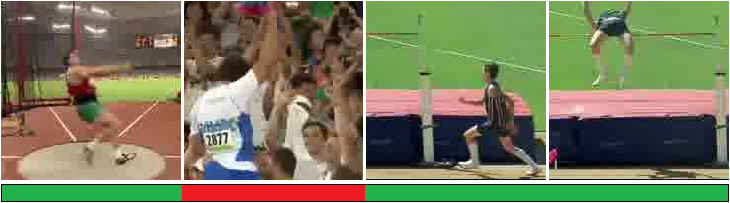}}
\end{center}
   \caption{Example of sports activities in a video which have to be localized temporally (green) and segments of non-sports activities that have not to be localized temporally (red).}
\label{fig:eyecatcher}
\end{figure}

Being well known, short videos containing only a single action are rather an artificial construct, being produced by special recordings of only a single action or by previously cutting the short video out of a larger one. More naturally, untrimmed videos usually feature no specific or limited length and contain more than a single action -- a video of the Summer Olympics may contain for example the sporting activities `high jump', `hammer throwing', and `fencing', but also non sporting parts, such as the commentators talking, interviews and shots of the crowd as well. Figure \ref{fig:eyecatcher} depicts an example. Another application field is the area of video surveillance, where often only a few time segments of large videos contain actions of interest, such as theft. Independent of the concrete application, time segments containing actions have to be identified from the whole video as accurate as possible in addition to the classification of the different actions taking place. As traditional approaches for solving this problem mostly use an expensive combination of sliding window mechanisms and classification, temporal action proposals generation was introduced as preprocessing step, searching for high-quality time segments first, which are thought to contain an action of interest with both high probability and good temporal localization. Thus, classification has to be performed only on the temporal action proposals.

A recent state of the art approach based on deep neural networks is the `Single-Stream Temporal Action Proposal' (SST) model \cite{sst}, processing videos utilizing 3D convolutions and a recurrent architecture. 
To the best of our knowledge, we are the first who investigate different positions and ways of fusion in two-stream architectures that utilize 3D convolutions on optical flow and image data for temporal action proposals generation. 
Our main contributions are: (1) The development of four two-stream model architectures for temporal action proposals generation originating from the SST model \cite{sst}. (2) Investigation and fine-tuning of the hyperparametrizations of the models. (3) Quantitative evaluation on the THUMOS'14 \cite{THUMOS14} dataset. (4) Showing the independence of a specific optical flow calculation method.  


\section{Related Work}

\textit{Action recognition} is a task to associate a single action class to a video. From this field, a lot of relevant innovation emerged. Two-stream convolutional neural networks \cite{simonyan2014two} were designed to process image data on the first stream and stacked optical flow fields on the second stream. The additional usage of stacked optical flow fields contributes temporal dynamical information of motion. Another approach was the extension of two-dimensional kernels used by the classical CNNs into the third dimension, therefore operating on 3D volumes defined by consecutive frames. The prominent C3D (Convolutional 3D) network \cite{c3d} employs this approach by processing videos divided into blocks of 16 consecutive frames. This is another way of utilizing temporal information. More recent approaches combine the two previous ideas: temporal information is utilized by applying 3D convolution on two streams, one using image data and the other one using optical flow. Among others \cite{khong2018improving, varol2018long}, the I3D (Inflated 3D ConvNet) network \cite{carreira2017quo} is a prominent example of that approach, coming to the conclusion that 3D convolutional neural networks also profit from a two-stream architecture. This insight in the field of action recognition serves in this work as inspiration to transfer that approach to the field of temporal action proposals generation.

The need for \textit{temporal action proposals} comes from the task of the temporal localization of actions in long, untrimmed videos and the classification of said actions. Before temporal action proposals, this problem was tackled with sliding window approaches: The extraction of overlapping time segments with varied length. Subsequently, a classification of each time segment was done to find the action in time. As this process was very time-consuming with a lot of time segments to be classified, temporal action proposals were invented to reduce the number of time segments that have to be classified. There exists early work \cite{caba2016fast} on temporal action proposals relying on traditional approaches. Among recent successful work \cite{escorcia2016daps,sst, gao2017cascaded,gao2017turn, lin2017temporal} it is instead common to take advantage of deep neural networks. Several works \cite{escorcia2016daps,sst, gao2017cascaded,gao2017turn} are utilizing 3D convolutional neural networks (3D ConvNets) for the generation of temporal action proposals -- an approach already known from the field of action recognition, see above. Being another prominent approach from the field of action recognition, two-stream networks with 2D kernels are used as well \cite{lin2017temporal,gao2017cascaded}, taking advantage of optical flow on the second stream. Despite being successfully used in action recognition, the combination of 3D convolutions with a two-stream network has not made it to common practice in the field of temporal action proposals generation yet.

In the field of \textit{temporal action localization} -- both the temporal localization and classification of actions in long, untrimmed videos -- the combination of 3D convolutions with two-stream networks found use recently in \cite{chao2018rethinking, nguyen2018weakly}. In most works, the temporal action proposal generation is a sub-task of the overall approach. However, there also exist end-to-end approaches \cite{buch2017end}.





\section{Methodic approach}
In this work, we follow the general approach presented by Buch \etal \cite{sst} for the SST model. Just like there, each video is divided into non-overlapping blocks of 16 frames and features are extracted using the C3D network \cite{c3d} as a feature extractor. Those features serve as input for a recurrent neural network, producing confidence scores for 32 possible time segments in each step. After post-processing with a score threshold and non-maxima suppression, a reduced set of temporal action proposals is generated. We stick to this approach and utilize the existing architectures, but extend them to a two-stream model architecture by introducing a second stream working on the corresponding images of optical flow, with the optical flow corresponding to image $j$ being calculated from image $j-1$ and $j$. Applying 3D convolutions on the optical flow allows efficiently making use of the dynamics of motion. We design four variants of this new architecture, differing in the position and way the separate streams are fused before continuing in a common stream. 

In the following, we will have a closer look at the four designed two-stream model architectures. All of them have in common that they process videos by dividing them into subsequent blocks of 16 images without overlap and processing them sequentially. For each block of 16 original images on the first stream -- called video stream -- there are the 16 corresponding images of optical flow on the second stream -- called flow stream. 
These blocks of 16 images are then first processed in parallel before the streams get fused later in the architecture. The position and the way of the fusion differ in the four model architectures. In the following, we will highlight the C3D network apart from the SST network which uses it as feature extractor.

\begin{figure*}
\begin{center}
\fbox{\includegraphics[width=1\textwidth]{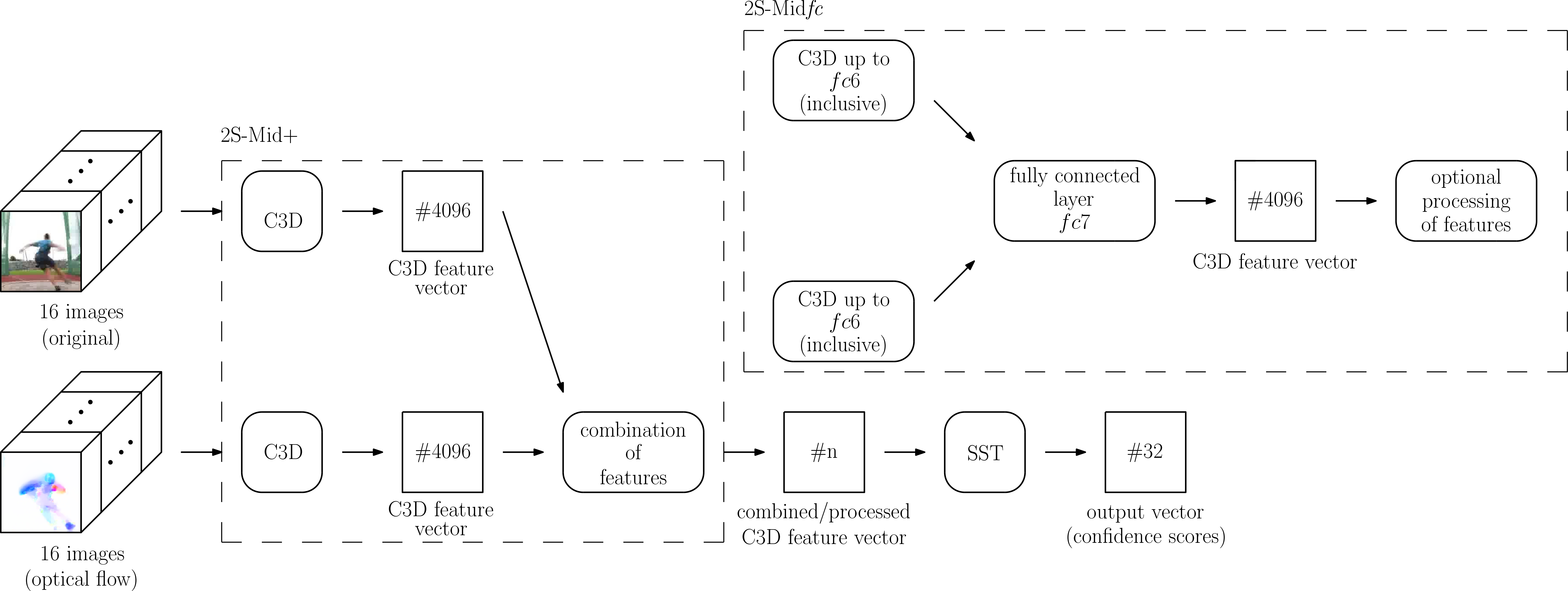}}
\end{center}
\caption{Outline of variant mid fusion by concatenation (2S-Mid+) and mid fusion by \textit{fc} layer (2S-Mid\textit{fc}) of the two-stream model architectures. To get 2S-Mid\textit{fc} from 2S-Mid+, simply replace the 2S-Mid+ content in the dashed box with the one of 2S-Mid\textit{fc}. For 2S-Mid+, the optional processing steps and the concatenation of feature vectors are bundled in the `combination of features'.}
\label{fig:var_1_and_2_scheme}
\end{figure*}

\paragraph{\#1: Mid fusion by concatenation (2S-Mid+)}
The first of the designed two-stream variants is fusing the two separate streams by concatenating features extracted by two separate C3D networks before being used as input into the SST network. This approach is inspired by Khong \etal \cite{khong2018improving} from the field of human action recognition. One of their investigated two-stream models utilizes C3D features extracted from the \textit{fc6}-layer of two separate C3D networks, one of them operating on the original images and the other one operating on optical flow. Among other processing steps, the two separate C3D feature vectors get concatenated there before being fed into a linear support vector machine (SVM) for classification.

The idea of fusing two streams by concatenating C3D features serves as a basis for our variant 2S-Mid+ of the designed two-stream networks. Two separate C3D networks get employed: one operating on the original images and one operating on the corresponding images of optical flow. The two streams stay separate until the end of the C3D networks where separate feature vectors are extracted. Optional processing of these feature vectors, like applying $L2$-normalization or principal component analysis, takes place after the extraction. The next performed step is the concatenation of the separate feature vectors. For block $i$ of a video, $f_{\text{\textit{v,i}}}$ denotes the feature vector from the video stream and $f_{\text{\textit{f,i}}}$ denotes the feature vector from the flow stream, which are concatenated and result in the concatenated feature vector $f_{\text{\textit{c,i}}}$. 
\begin{equation}
    f_{\text{\textit{c,i}}} = [f_{\text{\textit{v,i}}}^T, f_{\text{\textit{f,i}}}^T]^T 
\end{equation}
The concatenated feature vector $f_{\text{\textit{c,i}}}$ serves then as an input for the SST network, which determines confidence scores for temporal windows. A schematic representation of the resulting network is shown in Figure \ref{fig:var_1_and_2_scheme}.

Training: Just as with the original combination of C3D and SST network the C3D networks will be trained separately from the SST network on the task of action recognition. The SST network will be trained afterward based upon the extracted and combined C3D feature vectors of the pretrained C3D networks on the task of temporal action proposal generation.


\paragraph{\#2: Mid fusion by \textit{\textbf{fc}} layer (2S-Mid\textit{\textbf{fc}})}
\label{chap:var2_model}
The first variant 2S-Mid+ fuses the streams `by hand', as the fusion is not learned by the neural network but is performed by concatenation instead. A possible logical consequence is therefore to let the neural network learn how to fuse the two streams by combining the separate C3D networks in one of their later, fully connected layers, as it will be done in 2S-Mid\textit{fc}. This idea is supported by the work of Varol \etal \cite{varol2018long} on the field of action recognition, who use two separate C3D networks for original images and optical flow that are fused using a shared \textit{fc6} layer.

2S-Mid\textit{fc} uses -- just as 2S-Mid+ -- two separate C3D networks, one operating on the original images and one operating on the corresponding images of optical flow. In contrast, the two streams stay only separate up to the \textit{fc6} layers. For block $i$ of a video, $a_{\text{\textit{v-fc6,i}}}$ denotes the activations that are put out by the \textit{fc6} layer of the video stream and $a_{\text{\textit{f-fc6,i}}}$ denotes the activations that are put out by the \textit{fc6} layer of the flow stream accordingly. Both have $4096$ elements and serve as common input into the \textit{fc7} layer, thus delivering $8192$ elements. The shared \textit{fc7} layer fuses both streams, producing the output $a_{\text{\textit{c-fc7,i}}}$ with 4096 elements. In equation \ref{eq:var2_activation}, $R$ denotes the ReLU activation function.
\begin{equation}
    \label{eq:var2_activation}
    a_{\text{\textit{c-fc7,i}}} = R(W_{\text{\textit{fc7}}} \cdot [a_{\text{\textit{v-fc6,i}}}^T, a_{\text{\textit{f-fc6,i}}}^T]^T + b_{\text{\textit{fc7}}})
\end{equation}
The activation $a_{\text{\textit{c-fc7,i}}}$ is used as feature representation and optional post-processing can be applied before being used as input to the SST network. A schematic representation of the resulting network can be seen in Figure \ref{fig:var_1_and_2_scheme}.

\begin{figure*}[ht]
\begin{center}
\fbox{\includegraphics[width=1\textwidth]{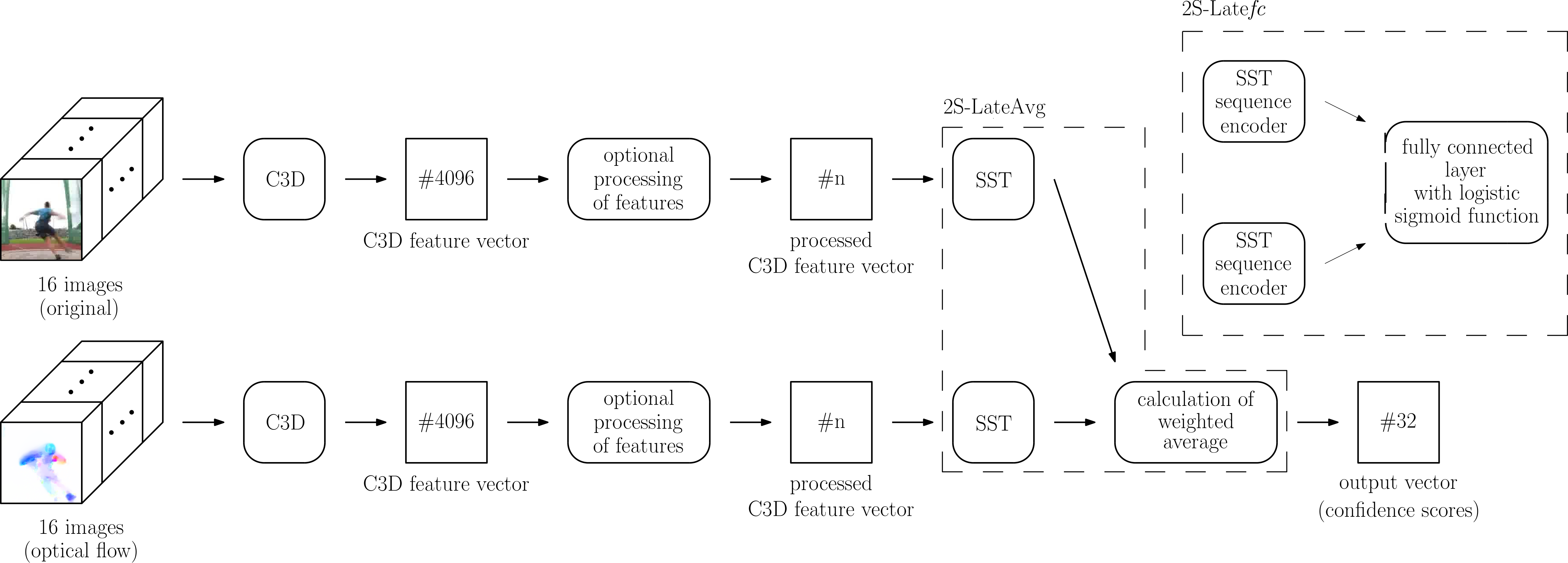}}
\end{center}
\caption{Outline of variant late fusion by weighted average (2S-LateAvg) and late fusion by fc layer (2S-Late\textit{fc}) of the two-stream model architectures. To get 2S-Late\textit{fc} from 2S-LateAvg, simply replace the 2S-LateAvg's content in the dashed box with the one of 2S-Late\textit{fc}.}
\label{fig:var_3_and_4_scheme}
\end{figure*}

Training: Two single-stream C3D networks are to be trained up front. The estimated weights are used to initialize the layers up to \textit{fc6}. As the dimension of the \textit{fc7} layer changed, it cannot be trained up front with a single-stream C3D network, so the two-stream C3D network with preinitialized weights up to \textit{fc6} has to be trained again on the task of action recognition. The network trained that way is then used to extract features, which are used to train the SST network on the task of temporal action proposals generation.


\paragraph{\#3: Late fusion by weighted average (2S-LateAvg)}



In the third variant 2S-LateAvg the fusion is moved to the very end of the network by forming a weighted average of two separate confidence score vectors. The idea is inspired by the temporal segment network (TSN) from Wang \etal \cite{tsn} for action recognition which fuses the separate streams by a weighted average of class scores.

2S-LateAvg utilizes two separate streams, each consisting of a full C3D and SST network. One stream operates on the original images, the other one on the corresponding images of optical flow. Both separate C3D networks extract separate C3D feature vectors, which are used as input into two separate SST networks. The SST networks are then used to generate separate vectors with confidence scores for the same time windows. For block $i$, the confidence score vectors of the video stream and the flow stream are called $c_{\text{\textit{v,i}}}$ and $c_{\text{\textit{f,i}}}$. The streams get fused by calculating the weighted average over these separate confidence scores with the weight factor $\alpha$, 0 $\leq$ $\alpha$ $\leq$ 1, and result in the common confidence score vector $c_{\text{\textit{c,i}}}$.
\begin{equation}
    c_{\text{\textit{c,i}}} = (1 - \alpha) \cdot c_{\text{\textit{v,i}}} + \alpha \cdot c_{\text{\textit{f,i}}}
\end{equation}
A schematic representation of the resulting network architecture can be seen in Figure \ref{fig:var_3_and_4_scheme}.

Training: The two separate C3D networks are pretrained on the task of action recognition. The two separate SST networks are then trained on basis of the extracted C3D feature vectors, one of them on C3D feature vectors extracted from the original images and the other one on C3D feature vectors extracted from images of optical flow. Training of the separate SST networks together based on the performance of the weighted average of the confidence score vectors is possible, but not mandatory. 


\paragraph{\#4: Late fusion by \textit{\textbf{fc}} layer (2S-Late\textit{\textbf{fc}})}
\label{chap:var4_architecture}
For 2S-LateAvg, the fusion of the separate streams is just as with 2S-Mid+ done `by hand', as the fusion is not learned by the network but done by calculating the weighted average over the confidence score vectors. Therefore, it seems logical to let the network learn how to fuse the two separate streams, which will be done in 2S-Late\textit{fc} by utilizing the fully connected layer at the end of the SST network. 2S-Mid\textit{fc}, where the second fully connected layer \textit{fc7} of the C3D network was used for the fusion, serves as inspiration. 

2S-Late\textit{fc} utilizes two separate C3D networks, one operating on the original images and one on the corresponding images of optical flow. Both are used to extract separate C3D feature vectors. They serve as input into two separate SST networks, one for the C3D feature vectors derived from the original images and one for the C3D feature vectors derived from the images of optical flow. Both SST networks stay separate until the end of the sequence encoders -- the recurrent part before the fully connected layer. The output vectors of the separate sequence encoders -- for block $i$ denoted as $s_{\text{\textit{v,i}}}$ for the video stream and $s_{\text{\textit{f,i}}}$ for the flow stream -- are used as input for a shared fully connected layer, which utilizes a logistic sigmoid function $\sigma$ to calculate the common confidence vector $c_{\text{\textit{c,i}}}$ in each step. 
\begin{equation}
    c_{\text{\textit{c,i}}} = \sigma(W_{\text{\textit{fc}}} \cdot [s_{\text{\textit{v,i}}}^T, s_{\text{\textit{f,i}}}^T]^T + b_{\text{\textit{fc}}})
\end{equation}
An outline of the resulting network is shown in Figure \ref{fig:var_3_and_4_scheme}.

Training: The separate C3D networks are to be pretrained just as in 2S-LateAvg, the same applies to the two separate SST networks. In contrast to 2S-LateAvg, the weights determined for the separate SST networks can only be used to initialize the two fused SST networks up to the end of the sequence encoder, as the dimension of the shared fully connected layer has changed. Therefore, the fused SST networks are to be trained again to calculate the weights for the fully connected layer, before they can be used for confidence score calculation.


\section{Evaluation}
In this section, a quantitative evaluation will be performed. First of all, we will investigate experiments regarding the hyperparametrization of the flow stream, followed by the evaluation of the four designed two-stream model architectures in comparison to the single-stream variants. The best configurations of fusion will be determined, as well as the improvement to the single-stream networks. Evaluation and training for temporal action proposal generation will be performed on the THUMOS'14 \cite{THUMOS14} dataset. The validation split will be used for the training as it is common practice on this dataset, while the test split remains for the evaluation. If the training of the C3D network is necessary, the UCF101 \cite{UCF101} dataset will be utilized. We are building upon a implementation\footnote{https://github.com/JaywongWang/SST-Tensorflow} of the SST network in TensorFlow, coming with already extracted features for the original video data of THUMOS'14. If not stated otherwise, the method of Brox \etal \cite{brox2004high} is used for optical flow calculation.

\subsection{Flow stream experiments}
\begin{figure*}
\begin{center}
\fbox{\includegraphics[width=0.46\textwidth]{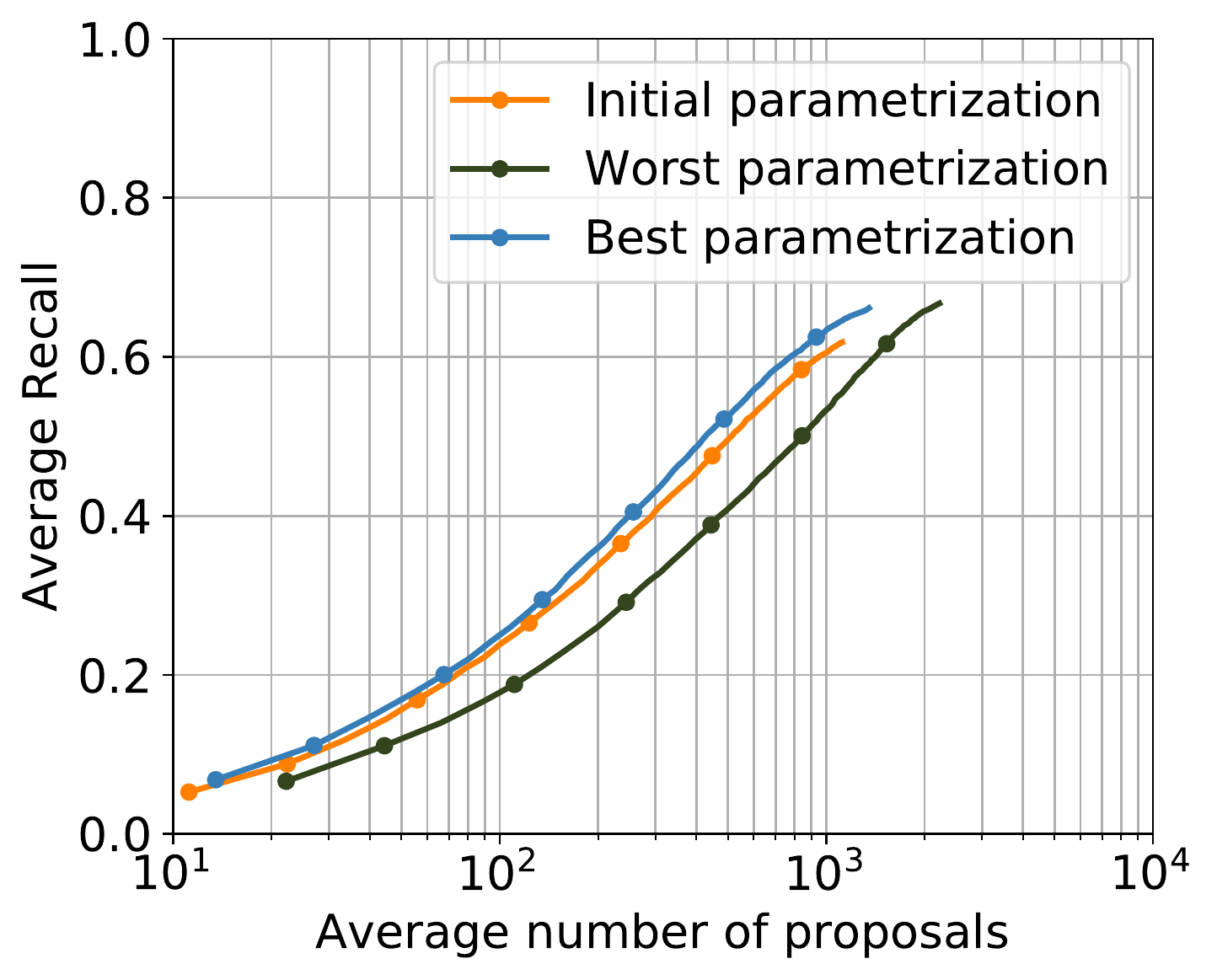}}
\begin{minipage}[t]{0.04\textwidth}\includegraphics[width=\textwidth]{pictures/dummy.png}\end{minipage}
\fbox{\includegraphics[width=0.46\textwidth]{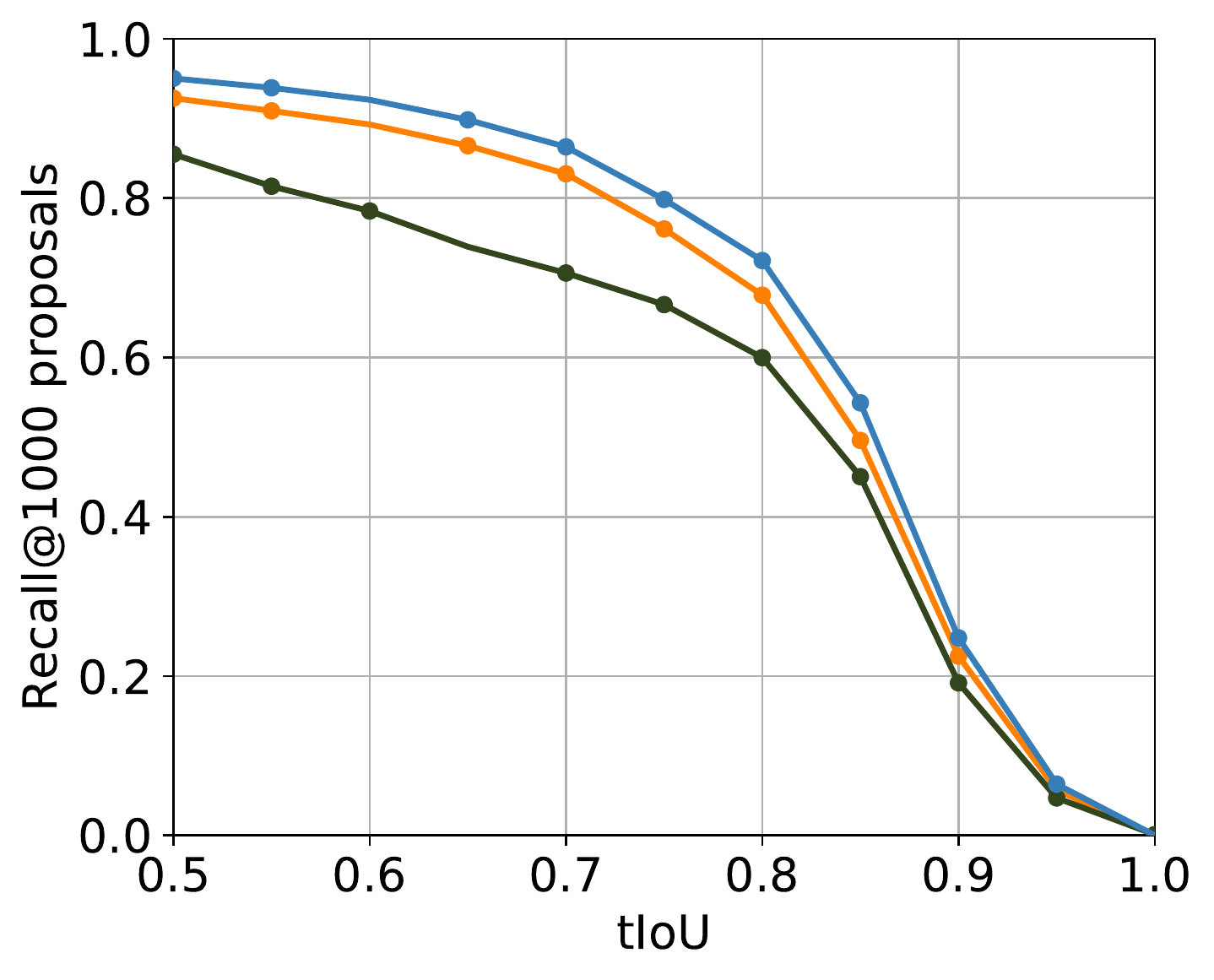}}
\end{center}
   \caption{Comparison of the results achieved with different parametrizations of the SST network using C3D features from images of optical flow as input. The high gap between best and worst parametrization results shows the importance of a proper parametrization, while the comparably small gap between initial and best parametrization indicates that the initial parametrization already worked well.}
\label{fig:flow_stream_res_comp}
\end{figure*}

As an initial step for the evaluation of the designed two-stream models, the hyperparametrization of the flow stream was investigated, working on images of optical flow. The parameters of the C3D network used for feature extraction remain untouched. For the SST network, several changes for parameters are investigated. C3D features from the \textit{fc6} layer of the C3D network are compared with features from the \textit{fc7} layer. One time the training of the C3D network is stopped early, the features from that C3D network are referred as early C3D features and compared with features from the C3D network when training is not stopped early; these features are referred as late C3D features. Two different preprocessing steps of the C3D features are investigated: \textit{L2}-normalization and principal component analysis for reducing the size of the feature vector from 4096 to 500 elements. Apart from these different inputs into the SST network, the parameters of the network itself are investigated: Different learning rates, different numbers of neurons per GRU layer, different numbers of GRU layers and different dropout rates. For the initial configuration, the parameter values of the video stream delivered with the used implementation are utilized. First, each parameter value got altered independently, afterward combinations of parameter changes on the basis of previous experiments were investigated. The initial parameter values, as well as the best configuration in the conducted experiments, can be seen in Table \ref{tab:param_flow_stream}. Not all parameter changes worked well, Figure \ref{fig:flow_stream_res_comp} shows the best results of the best configurations in comparison to the best results of the worst.

\begin{table}[h]
\begin{center}
\begin{tabular}{|l|c|c|}
\hline
Parameter & Initial Value & Best Value \\
\hline\hline
C3D features & late, \textit{fc7} & early, \textit{L2}, \textit{fc7}\\
Learning rate & 1e-3 & 1e-2\\
Dropout rate & 0.3 & 0.3\\
Neurons per GRU layer & 128 & 256\\
Number of GRU layers & 2 & 1\\

\hline
\end{tabular}
\end{center}
\caption{Configuration of the SST network operating on features from images of optical flow. The initial (left) and experimentally determined best (right) values are displayed. \textit{L2} stands for \textit{L2} feature normalization prior to usage.}
\label{tab:param_flow_stream}
\end{table}

\subsection{2S-Mid+ evaluation}
For these experiments, the already extracted features for the original images delivered with the TensorFlow implementation and the already extracted features for the images of optical flow extracted during the flow stream experiments are used. If being used, preprocessing steps are applied before those features are concatenated.

Experiments are conducted similar to the flow stream experiments, with the difference, that a reduced set of parameter values is explored based upon successful parameter values from the flow stream experiments. For each parameter setup, a new SST network is trained and evaluated on the concatenated features, as no pretrained model exists for the concatenated C3D feature vectors. Per configuration two SST networks were trained and evaluated. 

\begin{figure*}
\begin{center}
\fbox{\includegraphics[width=0.46\textwidth]{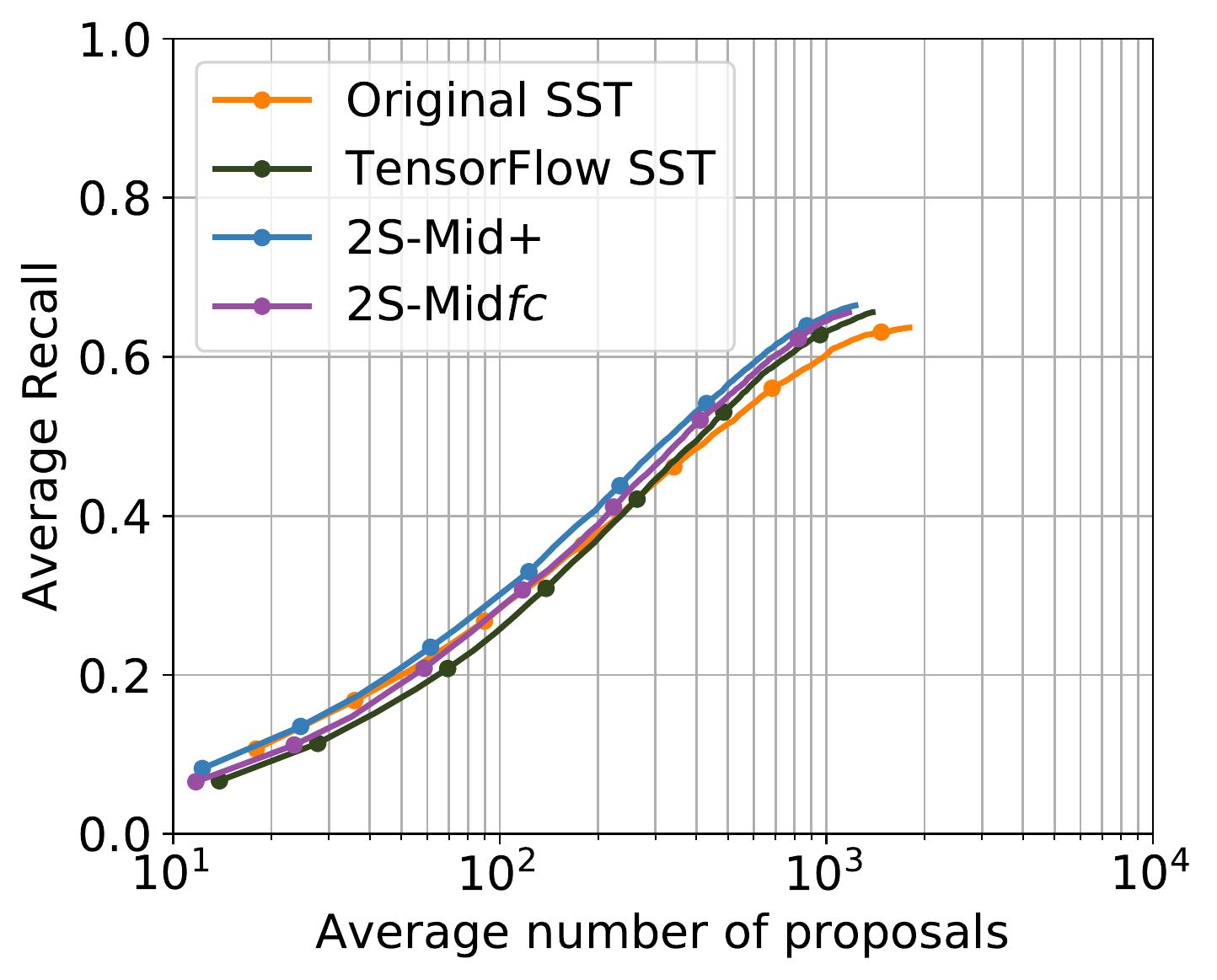}}
\begin{minipage}[t]{0.04\textwidth}\includegraphics[width=\textwidth]{pictures/dummy.png}\end{minipage}
\fbox{\includegraphics[width=0.46\textwidth]{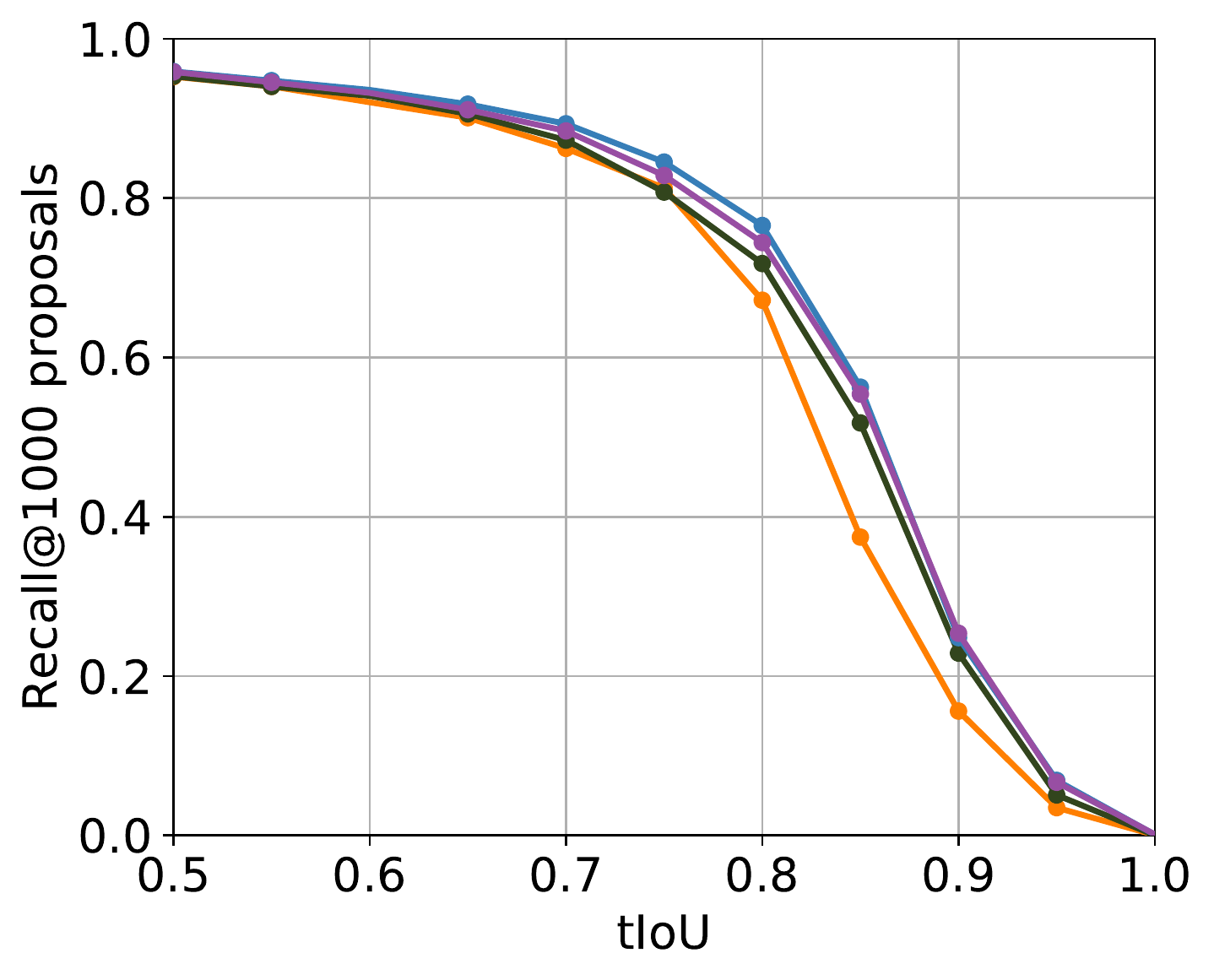}}
\end{center}
   \caption{Comparison between the results of 2S-Mid+ and 2S-Mid\textit{fc} of the two-stream model architectures using the experimentally determined best parametrization for them and the single-stream networks. For major parts of the metrics, the two-stream models manage to outperform the single-stream ones, with 2S-Mid+ outperforming 2S-Mid\textit{fc}.}
\label{fig:var_1_and_2_res_comp}
\end{figure*}

The best results were achieved with the configuration in Table \ref{tab:param_var1_stream}. In Figure \ref{fig:var_1_and_2_res_comp} the comparison with the single-stream networks -- the original SST network and the TensorFlow implementation of the SST network -- shows that the additional usage of optical flow leads to improvements for major parts of both metrics, while for minor parts in both metrics results of a comparable level were achieved.

\begin{table}
\begin{center}
\begin{tabular}{|l|c|c|}
\hline
Parameter & Shared Stream \\
\hline\hline
C3D features (images) & \textit{L2}, \textit{fc6}\\
C3D features (flow) & \textit{L2}, early, \textit{fc7}\\
Learning rate & 1e-2\\
Dropout rate & 0.3\\
Neurons per GRU layer & 256\\
Number of GRU layers & 2\\

\hline
\end{tabular}
\end{center}
\caption{Parameters and their experimentally determined best values for the SST network of 2S-Mid+ that operates on the concatenated feature vectors.
}
\label{tab:param_var1_stream}
\end{table}


\subsection{2S-Mid\textit{\textbf{fc}} evaluation}
The two streams are fused inside the feature extractor, thus, already extracted features cannot be used for the experiments. Instead, the existing weights for the feature extraction on image data and the weights determined during the flow stream experiments for feature extraction on images of optical flow are used for initialization. Training is done as described above using the UCF101 dataset for training the fused C3D networks. The weights used for initialization remain fixed in a first training phase used to determine the not preinitialized weights; an optional subsequent fine-tuning where no weights remain fixed is investigated as well. Solely the \textit{fc7} layer is investigated for feature extraction, as the streams are separate before that layer.


After the extraction of the C3D feature vectors, the training and subsequent evaluation of the SST network takes place. As with the experiments for the previous variant, different parameter configuration derived from successfull parameter values of the flow stream experiments are investigated, with two SST networks being trained and evaluated per configuration.

Among all experiments, the configuration in Table \ref{tab:param_var2_stream} produced the best results. The parametrization is, apart from the obvious deviation in the used C3D features caused by the design of the two-stream network, identical to the one which produced the best results for 2S-Mid+. The comparison of the performance concerning the two metrics can be found in Figure \ref{fig:var_1_and_2_res_comp}. Concerning both metrics, 2S-Mid\textit{fc} achieves in comparison with the single-stream networks improvements for major parts of both metrics as well but produces slightly worse results than 2S-Mid+.

\begin{table}
\begin{center}
\begin{tabular}{|l|c|c|}
\hline
Parameter & Shared Stream \\
\hline\hline
C3D features & \textit{L2}, no finetuning, \textit{fc7}\\
Learning rate & 1e-2\\
Dropout rate & 0.3\\
Neurons per GRU layer & 256\\
Number of GRU layers & 2\\

\hline
\end{tabular}
\end{center}
\caption{Parameters and their experimentally determined best values for the SST network of 2S-Mid\textit{fc}. 
}
\label{tab:param_var2_stream}
\end{table}


\subsection{2S-LateAvg evaluation}
Because fusion takes place right after the confidence scores of each stream were created, the pretrained SST network and the SST networks from the flow stream experiments can be used. For $\alpha$ the values 1/3, 1/2, and 2/3 are investigated.


Different sets of parameter values are explored for the hyperparameters of the SST network of the flow stream. The parameter values of hyperparameters of the SST network of the video stream remain untouched. To produce first results no further training is needed as the whole two-stream model can be initialized with pretrained weights, but an optional common fine-tuning of the two SST networks based upon the weighted average of the confidence scores is investigated as well.

\begin{figure*}
\begin{center}
\fbox{\includegraphics[width=0.46\textwidth]{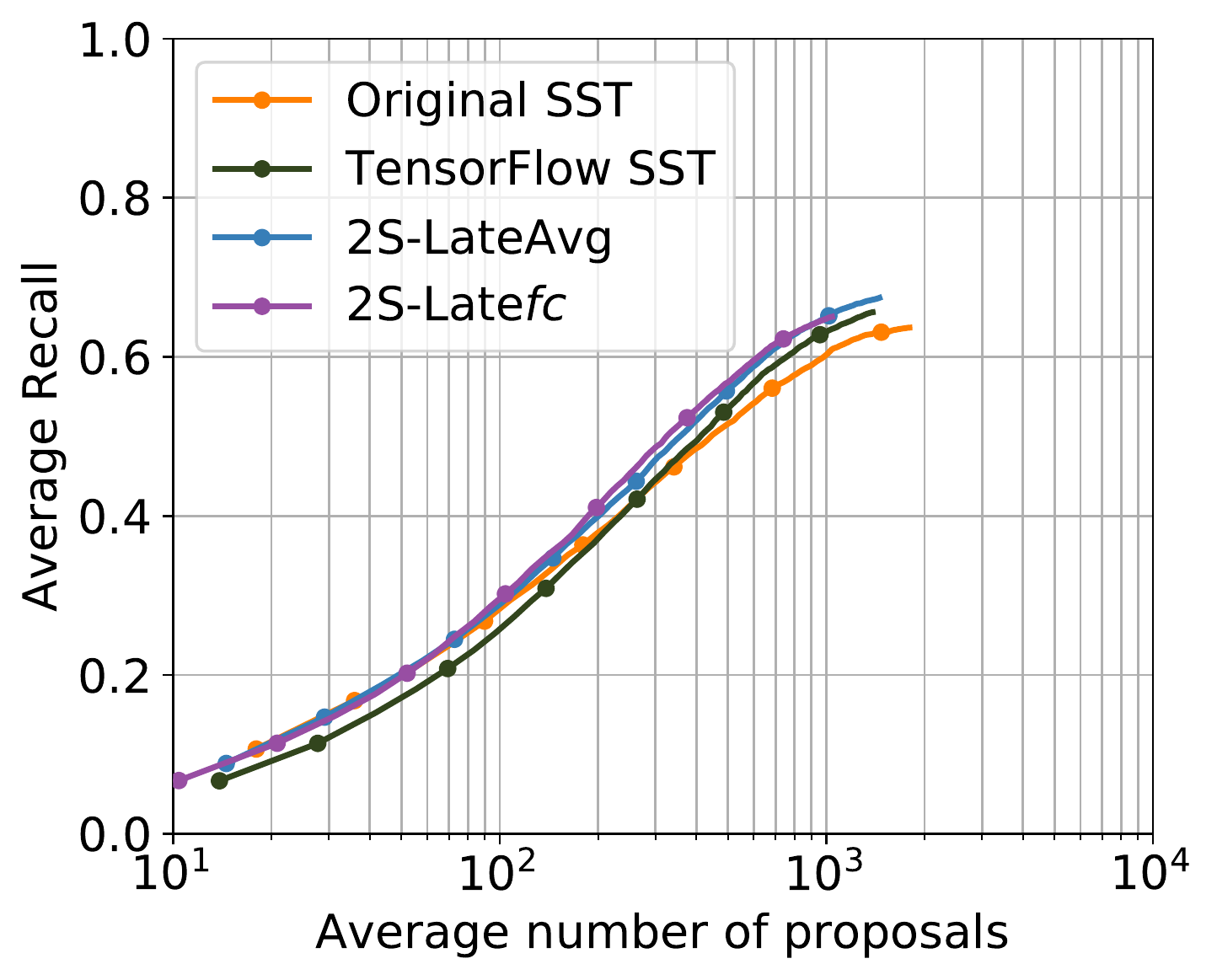}}
\begin{minipage}[t]{0.04\textwidth}\includegraphics[width=\textwidth]{pictures/dummy.png}\end{minipage}
\fbox{\includegraphics[width=0.46\textwidth]{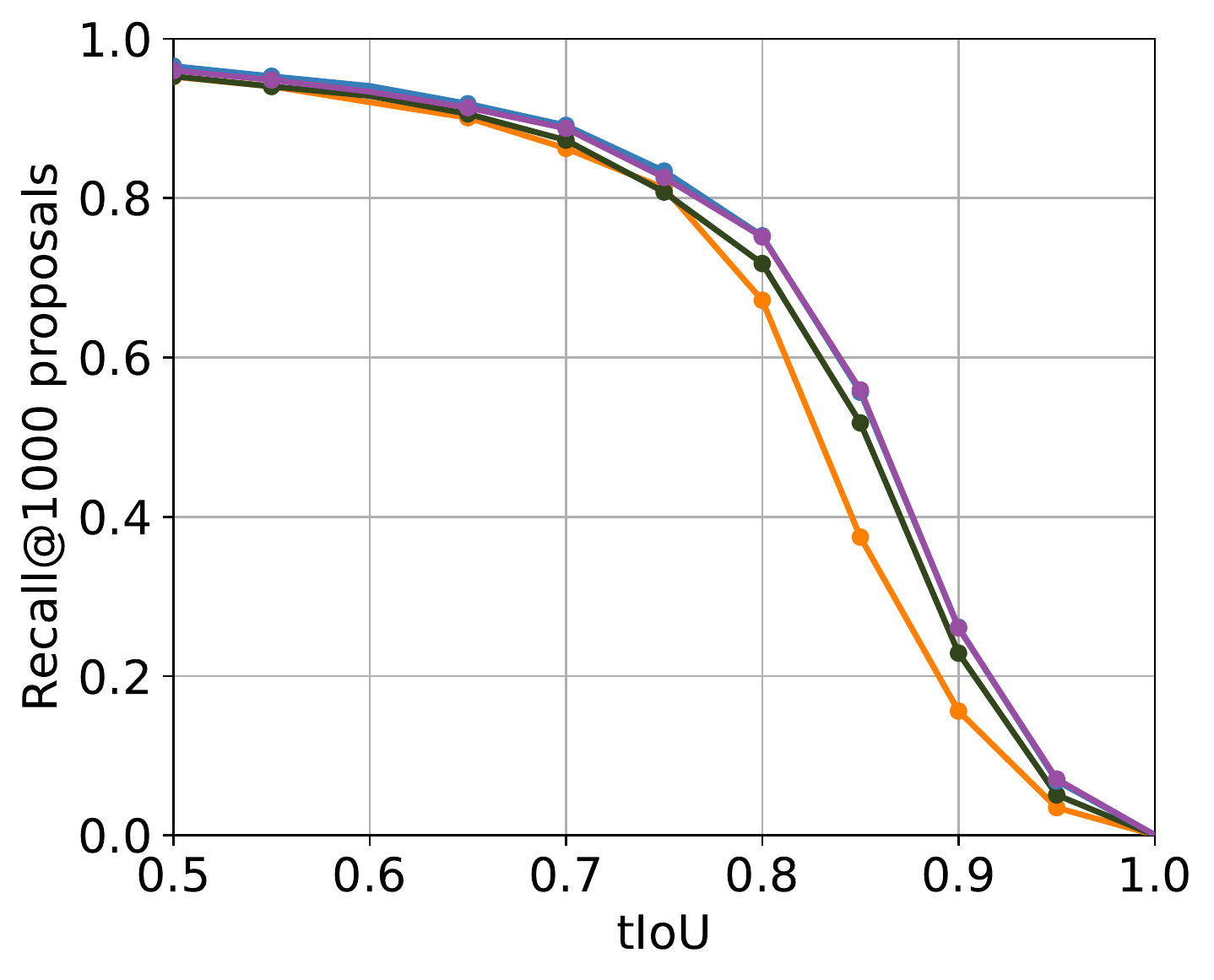}}
\end{center}
   \caption{Comparison between the results of 2S-LateAvg, 2S-Late\textit{fc} and the single-stream networks. For major parts of the metrics, the two-stream models manage to outperform the single-stream ones, with 2S-LateAvg also achieving improvements for very small tIoU.}
\label{fig:var_3_and_4_res_comp}
\end{figure*}

The parametrization that delivered the best results among all experiments for 2S-LateAvg is displayed in Table \ref{tab:param_var3_stream}. The comparison with the single-stream networks concerning the two known metrics is displayed in Figure \ref{fig:var_3_and_4_res_comp}. 
For major parts -- even for small tIoU -- improvements are achieved.

\begin{table}
\begin{center}
\begin{tabular}{|l|c|c|}
\hline
Parameter & Flow Stream & Image Stream \\
\hline\hline
C3D features & early, \textit{L2}, \textit{fc7} & \textit{fc6}\\
Learning rate & 1e-2 & 1e-3\\ 
Dropout rate & 0.3 & 0.3\\
Neurons per GRU layer & 256 & 128\\
Number of GRU layers & 1 & 2\\
Flow stream weight $\alpha$ & 0.5 & 0.5\\
\hline
Common Finetuning & \multicolumn{2}{c|}{Not used}\\

\hline
\end{tabular}
\end{center}
\caption{Parameters and their experimentally determined best values for the two separate SST networks utilized by 2S-LateAvg. Different parametrizations were only examined for SST network of the flow stream.
}
\label{tab:param_var3_stream}
\end{table}


\subsection{2S-Late\textit{\textbf{fc}} evaluation}
Feature extraction is performed as in 2S-LateAvg, but already trained SST networks cannot be employed totally, as the fusion is done by a shared fully connected layer of both separate SST networks. Therefore, weights of those networks can only be used for initialization up to the point of fusion. Training is done as described above. Weights used for the initialization remain fixed while training the fully connected layer, but an optional fine-tuning of all weights of the fused SST networks is investigated as well.


Similar to the experiments above the hyperparameters for the part belonging to the video stream remain fixed whereas for the flow stream they are explored. For each configuration two training and evaluation procedures are performed for the fused SST networks.

The parametrization producing the best results can be seen in Table \ref{tab:param_var4_stream}. It can be seen that the values of the parameters that are common to 2S-LateAvg and 2S-Late\textit{fc} are the same, therefore showing consistency. A comparison with the single-stream networks concerning both known metrics is shown in Figure \ref{fig:var_3_and_4_res_comp}. Again, improvements are achieved for major parts of both metrics in comparison with the single-stream networks.

\begin{table}
\begin{center}
\begin{tabular}{|l|c|c|}
\hline
Parameter & Flow Stream & Image Stream \\
\hline\hline
C3D features & early, \textit{L2}, \textit{fc7} & \textit{fc6}\\
Separate learning rate & 1e-2 & 1e-3\\
Common learning rate & 1e-3 & 1e-3\\
Dropout rate & 0.3 & 0.3\\
Neurons per GRU layer & 256 & 128\\
Number of GRU layers & 1 & 2\\
\hline
Common Finetuning & \multicolumn{2}{c|}{Not used}\\

\hline
\end{tabular}
\end{center}
\caption{Parameters and values for the two separate SST networks in 2S-Late\textit{fc}. The `separate learning rate' denotes the learning rate used to pretrain the two separate SST networks, whose weights are used to initialize the separate sequence encoders. The `common learning rate' denotes the learning rate used to train the common \textit{fc} layer after the preinitialized sequence encoders.}
\label{tab:param_var4_stream}
\end{table}


\subsection{Optical flow experiments}
Until now experiments were conducted using Brox \etal \cite{brox2004high} for optical flow. To investigate if the observed improvements can hold when the method of calculating optical flow is changed, for 2S-LateAvgFN FlowNet2 \cite{ilg2017flownet} is used. This method uses a neural network for supervised learning of optical flow in contrast to the traditional optimization approach of Brox \etal. 
A C3D network and a single-stream SST network are trained the same way as before, using the best configuration from 2S-LateAvg. The determined weights are used for initialization of the flow stream of 2S-LateAvgFN. Experiments with this parametrization and these weights are conducted just as when using optical flow calculated with Brox \etal. The results are slightly worse compared to the case where Brox is used, but remain on a comparable level, achieving improvements in comparison to the single-stream networks. 


\subsection{Summary}
All four two-stream models lead to improvements compared to the single-stream networks. This indicates that the utilization 3D convolutions in a two-stream setup makes sense for the task of temporal action proposal generation. A tabular comparison is shown in Table \ref{tab:res_summary}. 2S-Mid+ and 2S-LateAvg perform best, with negligible differences in performance. They have in common that the fusion of both streams takes place outside of the actual neural networks, thus does not get learned. 

\section{Conclusion}
In this work, four different two-stream model architectures with different fusions utilizing sequences of images on one stream and images of optical flow on the other stream were investigated for the purpose of temporal action proposal generation. Utilizing sequences of images of optical flow on the second stream in addition to sequences of the original images on the first and processing them using 3D convolutions on both streams, improvements where achieved for all explored two-stream models in comparison to the single-stream models omitting a second stream. It was also shown that the improvement is not bound to using a certain method of calculating optical flow by investigating another one and achieving improvements as well. Apart from showing that the general approach of combining a two-stream architecture with 3D convolutions is beneficial for the task of temporal action proposal generation, a suitable basis for further work on the larger field of action localization has been created.

\begin{table}[hb]
\begin{center}
\begin{tabular}{|l|c|c|}
\hline
Network & Score\\
\hline\hline
Original SST network & 0.6025\\
TensorFlow Implementation of SST network & 0.6295\\
SST network (images of optical flow) & 0.6320\\
2S-Mid+ & 0.6497\\
2S-Mid\textit{fc} & 0.6438\\
2S-LateAvg & 0.6495\\
2S-Late\textit{fc} & 0.6466\\
2S-LateAvgFN & 0.6436\\
\hline
\end{tabular}
\end{center}
\caption{Comparison of the single-stream networks with the different two-stream models. The displayed score refers to the metric `average recall at average 1000 proposals'. The scores for the two-stream networks and the single-stream network with optical flow come from the best experiments presented in this work. It can be seen that all the single-stream variants of the SST networks are surpassed by every single two-stream model, even if the calculation method of optical flow is changed. Best results are achieved with 2S-Mid+ and 2S-LateAvg.}
\label{tab:res_summary}
\end{table}

{\small
\bibliographystyle{ieee}
\bibliography{egbib}
}

\end{document}